# AI SURVIVAL STORIES: A TAXONOMIC ANALYSIS OF AI EXISTENTIAL RISK

*Herman Cappelen*[a], *Simon Goldstein*[b,c] , *John Hawthorne*[d]

[a] Department of Philosophy, The University of Hong Kong, HK,
[b] Australian Catholic University, Melbourne, AU,
[c] Lingnan University, Dept of Philosophy, Tuen Mun, HK,
[d] Department of Philosophy, University of Southern California, US

*Abstract:* Since the release of ChatGPT, there has been a lot of debate about whether AI systems pose an existential risk to humanity. This paper develops a general framework for thinking about the existential risk of AI systems. We analyze a two-premise argument that AI systems pose a threat to humanity. Premise one: AI systems will become extremely powerful. Premise two: if AI systems become extremely powerful, they will destroy humanity. We use these two premises to construct a taxonomy of 'survival stories,' in which humanity survives into the far future. In each survival story, one of the two premises fails. Either scientific barriers prevent AI systems from becoming extremely powerful; or humanity bans research into AI systems, thereby preventing them from becoming extremely powerful; or extremely powerful AI systems do not destroy humanity, because their goals prevent them from doing so; or extremely powerful AI systems do not destroy humanity, because we can reliably detect and disable systems that have the goal of doing so. We argue that different survival stories face different challenges. We also argue that different survival stories motivate different responses to the threats from AI. Finally, we use our taxonomy to produce rough estimates of 'P(doom),' the probability that humanity will be destroyed by AI.

## 1. Introduction

In the last few years, AI systems have improved dramatically in capabilities. The widespread adoption of AI technology has drawn attention to the possibility that AI systems could pose an existential threat to humanity. But the precise nature of this threat remains contested.[1]

Why think that AI is an existential threat to humanity? Our analysis of AI risk is anchored around a two-premise argument that AIs are an existential threat to humanity. Premise one says that AI systems will continue to improve their capabilities until they become extremely powerful. Premise two says that if AI systems do become extremely powerful, they will go on to destroy humanity.

To analyze this risk, we introduce a taxonomy of the most likely 'survival stories,' in which humanity survives the existential threat of AI. Each survival story highlights one of the ways that the future could go if humanity were to survive rather than be destroyed by AI.[2] We'll critically analyze four survival stories. The first two survival stories involve *plateaus*, where something prevents AI systems from developing. This leads to a failure of premise one:

> **Technical Plateau:** Scientific barriers prevent AI systems from becoming extremely powerful.
>
> **Cultural Plateau:** Humanity bans research into AI systems becoming extremely powerful.

---

1 For a survey of skeptical attitudes to AI risk see Ambartsoumean & Yampolskiy(2023).

In the second two survival stories, premise two fails: extremely powerful AI systems emerge, but do not destroy humanity.

> **Alignment:** Extremely powerful AI systems do not destroy humanity, because their goals prevent them from doing so.
>
> **Oversight:** Extremely powerful AI systems do not destroy humanity, because we can reliably detect and disable systems that have the goal of doing so.

We argue that these four survival stories are the four main paths humanity may take to avert destruction from AI. The model of risk is depicted in Figure 1:

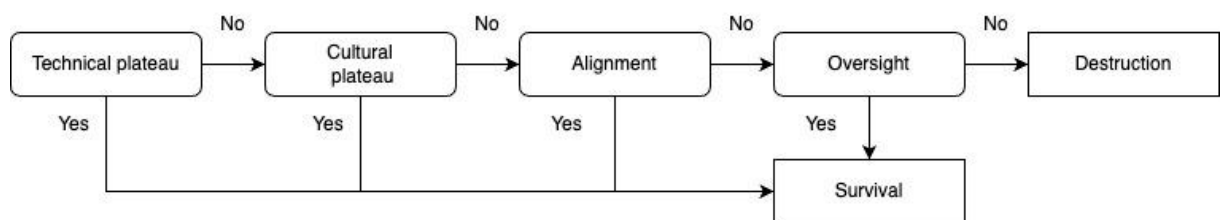

**Figure 1**
A Swiss cheese model of AI risk.

This is a 'Swiss cheese' model of accident prevention.[3] We have four 'layers' of safety. As long as one layer activates, humanity will survive. Our prospects for survival thus depend on how reliable each layer of safety is, and on whether the 'holes' in each layer of safety are independent of the others.

There is a growing literature on AI risk. Much of this literature emphasizes stories of destruction rather than of survival. Hendrycks et al. (2023) taxonomize the different risks of AI systems, showing how misuse, misalignment, and race dynamics could lead to AI systems destroying humanity. In an alternate framework, Carlsmith (2022) and Dung (2024) develop a series of premises that imply that AI is an existential risk.[4]

When the analysis focuses on stories of destruction rather than of survival, the natural instinct is to point out the speculative nature of the relevant destruction story. Our analysis differs from existing treatments in its emphasis on stories of survival rather than destruction.[5] This approach offers several key advantages over existing treatments.

---

[2] For humanity to survive into the far future, we'll also need to survive other threats besides AI, such as nuclear war and pandemics. See Posner (2004) or Ord (2020) for general overviews of catastrophic risk. Our focus will be on AI risk in particular, but we return to this point in our calculation of P(doom).

[3] See https://en.wikipedia.org/wiki/Swiss_cheese_model.

[4] Each premise in Dung (2024) can be negated to create a survival story. In this way, our approach will broadly complement the argumentation in that excellent paper. As we explain below, our paper extends the argumentation from Dung (2024) in the details of the challenges we raise for each survival story, especially cultural plateau and oversight, in our analysis of how different survival stories recommend different policy responses, and in our estimation of P(doom).

[5] Our analysis will not focus on exactly how AI systems cause the destruction of humanity. For example, we will not consider the relative likelihood of destruction through military weapons, or engineering a deadly virus, or gradual economic disempowerment, or through some other means.

First, the focus on survival stories shifts the framing from justifying AI risk to examining specific paths to survival. This reframing puts the burden of proof on risk-skeptics to explain how humanity will survive, rather than merely criticizing risk arguments. It makes clear that those who reject AI-skepticism rely on implicit, unexamined assumptions about the future of AI. While destruction stories may at first glance appear speculative, it turns out that survival stories may suffer from the same vices.

In particular, sections 2 and 3 argue that each survival story faces its own challenges, which are structurally independent of challenges to other stories. The challenge to technical plateau is that the recent development of AI capabilities, along with the talent and money devoted to improving AI, suggest that AI systems will continue to develop until they are extremely powerful. The challenge to the cultural plateau is that AI development is a race that no one player can stop: a ban on capability-improving AI research would require the whole world to collectively agree, despite each actor's interest in reaping the benefits of advanced AI systems. The alignment challenge is that we should expect the goals of powerful AI systems to conflict with our own, given both resource scarcity and humanity's plans to control AI. The challenge to oversight is that arbitrarily powerful AI systems require arbitrarily reliable safety mechanisms, but there are bottlenecks to how much reliability we can achieve. AI optimists need to respond to these challenges story by story.

Second, our taxonomy of survival stories offers insights into future pathways toward productive safety research. In particular, section 4 argues that each survival story motivates a distinctive strategy for lowering AI risk. Considering survival stories one by one helps clarify the risks of AI systems. It also helps clarify the optimal strategies for making AI systems safer. We argue that different survival stories motivate different responses to the threats from AI. Technical plateau stories require very little of us. Cultural plateau survival stories suggest that we should try to ensure that when AI accidents happen, we leverage them to ban the development of more powerful AI systems. Alignment stories suggest that we should try to cooperate with AI, creating conditions under which they are happy to coexist with us. Perfect safety requires that we try very hard to prevent accidents and effectively monitor potentially misaligned AI systems. Differing attitudes towards the relative plausibility of different survival stories should therefore motivate different plans for mitigating risk.

Finally, our taxonomy of survival stories offers a method for estimating P(doom), the chance that humanity is destroyed by AI systems. In general, one productive way to estimate the probability of an event is to split it into a series of sub-events. Our taxonomy allows us to split survival into a series of survival stories. Our analysis of each survival story offers a constructive way to begin to estimate the overall chance that humanity survives. We turn to this task in section 5. We argue that in order to rule out any significant AI risk, the AI optimist would have to be unreasonably confident in their survival stories.

Before going on, we will clarify what we mean by 'surviving' an 'existential risk' from 'extremely powerful' AI systems. First, we will focus on the time frame of a few thousand years (although the challenges we raise could also be applied to shorter time frames). Second, it is worth distinguishing a few concepts of existential risk. First, there is extinction, in which human life is completely eliminated. Second, there is near-extinction, in which the human population is drastically reduced over thousands of years. Third, there is a loss of autonomy, in which AI systems become so dominant that humans lose the ability to make meaningful choices and control their own destiny. Fourth, there is a loss of control, in which AI systems have political control over humanity, but may still allow humans a good deal of autonomy in their personal lives.

In this paper, we'll use the term 'existential risk' synonymously with the phrase 'risk that humanity does not survive'. But we will not interpret this to require literal extinction. Rather, we will use it loosely to cover all three: extinction, near-extinction, and loss of autonomy. But we will not use it to include mere loss of control. Finally, by 'extremely powerful' AI systems, we have in mind systems that have the kind of capabilities required to threaten humanity's survival, as defined above. We'll discuss such capabilities in the next section when considering technical plateau. Nothing in this choice of terminology will substantively impact the arguments we make (though of course the stricter the notion of doom, the lower its probability); the reader is free to substitute their own terms.[6]

## 2. Plateau Survival Stories

Our taxonomy divides survival stories into two kinds: plateau and non-plateau. With plateaus, premise one fails, so that AI systems never become capable enough to pose an existential risk to humanity. With non-plateau survival stories, AI systems do become arbitrarily capable, and yet still do not destroy humanity, so premise two fails.

We'll begin our taxonomy with plateau stories. There are two different kinds of plateaus. In technical plateau, scientific barriers stop AIs from continuing to improve in capability. In cultural plateau, humanity collectively bans the development of any AI system powerful enough to destroy humanity. Let's consider each story in turn.

### 2.1. Technical Plateau

In the story of technical plateau, researchers never figure out how to make AI systems that are powerful enough to pose an existential threat to humanity. The reason they fail is that the science is too difficult.

In support of technical plateau, one tempting thought is that superintelligence is just impossible, or even incoherent. One such challenge is discussed in Chalmers (2010) maybe we're not able to compare intelligence levels between systems, such that one system is more intelligent than another, or is "superintelligent" compared to humans. Intelligence may not be a unitary, quantifiable property. Intelligence involves many different cognitive capacities, including learning, reasoning, problem-solving, language use, and creativity. It's not clear that these capacities can all be subsumed into a single general property that can have a well-defined magnitude allowing comparisons of "greater" or "lesser" intelligence. Even if we grant there is such a general property, it may not "add up" in a way that would straightforwardly scale. Incremental improvements to different cognitive capacities may not lead to an overall "intelligence" that is greater in the singular sense needed. So while we can make AIs that exceed humans on particular well-defined tasks, this may not lead to a general "superintelligence" that is greater than humans in an overall, singular sense. The notion of an increasing sequence of AIs may not be coherent. (On the other hand, Bostrom (2014) observes that we can make sense of superintelligence in terms of *speed*, imagining a being that can perform computations a trillion times faster than a human, but otherwise has the same thoughts.)

We will raise a few challenges to technical plateau: (i) recursive self-improvement has the potential to produce superintelligent AI systems; (ii) even without superintelligence, AI systems with roughly human-level intelligence could pose an existential threat to humanity; and (iii) there are good reasons to think that such AI systems will soon be developed. Let's consider each point in turn.

First, why think there could be superintelligent AI systems? One reason comes from 'recursive self-improvement'. Once AI systems reach a roughly human level of intelligence, it will be possible to use AI systems to improve the intelligence of AI systems. But in this case, each improvement in the intelligence of AI systems will also increase the rate at which the intelligence of AI systems can be improved.[7] There is a risk that this could lead

---

to exponential growth in intelligence. In fact, recursive self-improvement is an active research goal in machine learning.[8]

Second, the existential threat of AI may not require the science fiction of superintelligence. Another route to very powerful systems involves 'supernumerosity' rather than superintelligence. Imagine that in the next few decades, AI labs succeed in developing artificial general intelligence, or AGI. Imagine a future in which millions of human-level AIs are placed in control of large swaths of the economy and weapons sys-tems. Strategic competition between humans and these AIs could ultimately lead to de-struction.

Third, there are good empirical reasons to think that we are well on our way to AGI. First, many of the impressive results of contemporary LLMs have been predicted by 'scaling laws', which suggest that improvements in accuracy follow predictably from increased investment in computing power, measured in terms of floating point operations (Kaplan et al., 2020). Second, there is evidence that as we train models with more compute, they also tend to nonlinearly develop new capabilities. In this way, we might expect that massive increases in compute could lead to new LLMs that approach near-perfect accuracy, and develop more agential abilities, such as the development of world models I LDRIM and long-term planning.

On the other hand, defenders of technical plateau sometimes argue that scaling large language models is unlikely to produce AGI, because planning abilities cannot come from s caling.[9] According to this view, developing AGI will require fundamental architectural innovation. Another concern is that there is not enough human data to train new generations of LLMs, and we should expect LLMs to fail to learn from synthetic data (Shumailov et al., 2023). Still, these responses may be more relevant to how AI progresses over the next f ew decades rather than millennia.

Finally, it is worth clarifying the scope of technical plateau. There are many different routes by which extremely powerful AI systems could lead to the demise of humanity (see Hendrycks et al., 2023) for a survey of these different risks). Two of the most salient routes are misuse by a terrorist group and rogue misaligned AI. We interpret technical plateau s trongly so that it rules out AI systems that are powerful enough to destroy humanity by either of these routes.[10]

## 2.2. Cultural Plateau

In cultural plateau stories, humanity collectively implements a ban on capability-improving AI research. This ban blocks AI systems from becoming powerful enough to pose an existential threat.

Here, one definitional question is whether stories of cultural plateau are compatible with stories of technical plateau. Imagine that there are scientific barriers to building powerful AI systems, but humanity bans such research anyway, not knowing it would be unsuccessful. It won't ultimately matter for our paper whether this is counted as a case of 'technical plateau' versus 'cultural plateau'. As we show in section 5, we can estimate the overall probability of survival in a way that bypasses this terminological distinction. In particular, to estimate the overall chance of survival, we will suppose that each previous layer of safety has failed, and see how likely the next layer of safety is to succeed. We will do this through each of our four layers of safety, until we have calculated the overall probability of survival.

---

[8] For a list of recent examples of AIs improving AIs, see https://ai-improving-ai.safe.ai/.
[9] Gary Marcus is one prominent example of this line: see for example Marcus (2024). Also see Dreyfus (1978); Lake et al., (2017), Marcus (2018) for more general criticism.
[10] An alternative interpretation of some misuse scenarios would be one in which humanity is destroyed by another cause besides AI. This interpretation would be more salient if the relevant AI system were not very powerful, but marginally increased the capabilities of a very powerful terrorist organization so that it became powerful enough to threaten humanity. The exact taxonomic classification of these cases won't make a very strong difference for our analysis, for example because our later model of P(doom) can be accommodated to incorporate catastrophic risks besides AI.

In practice, this means that we may as well suppose that previous layers fail when considering the next layer of safety.[11] [12]

To think through cultural plateau, we'll proceed in two steps. First, we'll characterize what a cultural plateau looks like. Then, we'll consider three challenges for bringing about a cultural plateau: (i) it would be difficult for humanity to agree that AI is an existential threat; (ii) even then, many individual actors have powerful incentives to continue making AI systems; (iii) banning AI development is a collective problem, because the relevant actors are engaged in a race.

What would bans on capability research look like? We are looking for the kind of ban that would protect humanity from advanced AI systems for thousands of years. So the relevant ban needs to be an equilibrium. There are a few potential equilibria to consider. First, we might try to keep fairly powerful AI systems, but ban any research into more powerful AI systems. Perhaps current systems are allowed, but AGI is not. Here, the concern is that the line will be fuzzy. A more radical plateau would ban machine learning altogether. The idea here is that optimization algorithms are the fundamental technology leading to danger, and those algorithms are present whenever there is machine learning. An even more radical plateau would ban artificial computation, for example by outlawing programming languages.

The cultural plateau story also needs to specify how bans would be enforced. One thought is that the governments of the world could multilaterally coordinate to ban AI research. Such a ban might look like international agreements to ban chemical weapons. Another enforcement technique would be more broadly cultural. In one hopeful scenario, strong taboos emerge against AI capability research that are analogous to those against human cloning and eugenics.

Another important aspect of ban implementation will concern chips. Large AI models are trained on special computer chips that are both scarce and extremely difficult to manufacture. There are promising proposals for chip monitoring that could ensure that any such chip sold around the world could be restricted in its usage (Shavit, 2023). In a more radical implementation, there could be a ban on the production of any chip that could be used to train a sufficiently powerful model. Careful bans on powerful chips could allow for relatively safe AI research to continue while limiting the kind of AI research that would lead to extremely powerful systems.

We're now ready to consider the prospects for bans on AI research. We'll begin with a few challenges. The first challenge is the lack of global consensus that AI systems are an existential threat. On this point, here are the words of Anthropic's CEO, Dario Amodei, in a recent interview:

> "…again, I get to this idea of demonstrating specific risk. If you look at times in history, like World War I or World War II, industries' will can be bent towards the state. They can be gotten to do things that aren't necessarily profitable in the short-term because they understand that there's an emergency. Right now, we don't have an emergency. We just have a line on a graph that weirdos like me believe in and a few people like you who are interviewing me may somewhat believe in. We don't have clear and present danger."[13]

---

11 As an analogy, imagine that you're stranded on a desert island. Your survival depends on either being picked up by a boat from company 1 or a boat from company 2. Each company has a chance of sending a boat to your island, and a chance of stranding you. You can only get on one boat. Imagine that in fact both companies send a boat to you, and you get on the company 1 boat. Here, we can ask whether this is a 'company 1 survival story' or a 'company 2 survival story'. We could define these terms in one of two ways. In the narrow sense, you survived due to company 1, because that is the boat you took. But in a broad sense, you also survived due to company 2, because both companies sent a boat. When estimating the chance of survival, though, the natural question would be the chance that company 2 sends a boat even if company 1 doesn't. In this way, it doesn't matter whether we define survival stories in the narrow or broad sense, because in practice the narrow sense will be the relevant term for probability calculation.

12 A related question is how exactly to draw the line between technical and cultural plateau. Scientific development of AI systems does not merely require that various abstract facts about algorithms are true; rather, science itself requires the right kinds of investment, talent, and norms. In this way, scientific progress is itself cultural. Imagine that scientific progress into AI struggles because of lack of Without agreement that highly capable AI systems pose a great risk to humanity, it is hard to see why humanity would ban them.

The second challenge is that there is a great temptation for governments and labs to develop highly capable AI systems. AI systems have many tempting commercial applications. Machine learning engineers are intrinsically motivated to develop more powerful AI systems.

A related problem here is *points of no return*. As AI systems become more powerful, we will integrate them deeply into our personal lives, public culture, and economic structures. This will make it harder and harder to resist further AI development.

Of course, the benefits for individual parties of developing AI will be weighed against their costs: governments and labs are also motivated not to destroy humanity. But this brings us to the third challenge: collective action. Even if all of the relevant parties agreed that highly capable AI systems are too risky, they may fail to agree to ban them. The problem is that AI development is a race. Each party trying to develop AI systems has the potential to reap vast rewards if they succeed. They do not fully internalize the risks that they create in their efforts. Rather, the risks are born by everyone; and even if one party stops developing AI, their rivals will continue to create risk.

In the face of these challenges, what would it take to produce a ban? In any such story (i) relevant decision-makers must be convinced that AI is extremely dangerous, to the point that (ii) the risks of AI outweigh the benefits of developing AI, and in addition (iii) those decision-makers must be able to effectively coordinate to overcome collective action prob-lems.

We see at least three paths to this outcome. The first path is *sophisticated reflection*. In this story, humanity comes to accept on the basis of reflection on the nature of AI systems that it is simply too dangerous to develop extremely powerful AIs. After significant public debate, the major world governments agree to implement a ban. Here, one precedent might be humanity's success in ending chattel slavery. We can imagine two versions of sophisticated reflection. One story is *moral*: the key decision-makers become convinced that AI is so risky that it would be unethical to allow its development. Another story is *self-interested*: the key decision-makers realize that developing ever more powerful AI systems could disrupt the status quo that favors them, and they ban future AI development as a means to entrench their own authority.

The second path is *AI persuasion*. In this story, humanity does not itself have the wisdom to ban AI research. But our AI creations do. For example, Salib (2023) argues that AI systems will themselves create a plateau because they will not want to self-improve. In this story, future AI systems use powers of persuasion to convince humanity to ban the research. In this story, the amazing powers of future AI systems overcome the three challenges we mentioned.

The third path is *accidents*. In this story, the ban on AI research is caused by AI accidents, or "warning shots". Accidents could help overcome the first challenge to cultural plateau. After enough accidents take place, humanity might collectively become convinced that AI systems are too dangerous to develop. Accidents could also help with the third challenge, producing "Schelling points" in which the relevant parties are able to effectively coordinate to reach an agreement to ban capability-improving research. For example, an accident might make salient to all parties the possibility of banning the specific capability that produced the accident. What kind of accidents are likely to causally promote bans? Several features are relevant. First, size.

A tempting thought is that as accidents increase in size, it becomes more likely

---

funding, and because of cultural norms inside the ML community. We could think of this as either a case of technical or of cultural plateau. We leave to the reader the decision of where exactly to classify individual cases. What matters for us, though, is that the key *challenges* we raise to technical and cultural plateau will still affect these edge cases.
13 See https://www.nytimes.com/2024/04/12/podcasts/transcript-ezra-klein-interviews-dario-amodei.html.

for them to create a cultural plateau. For example, if an AI system wiped out half of humanity, surely the survivors would agree to permanently ban the use of AI systems. On the other hand, it requires too much trust in humanity to imagine a very tiny accident convincing humanity to stop AI research. But the larger the size of the accident, the less likely it is that the accident will take place. So there is a spectrum of accidents: small accidents are likely to happen, but not likely to cause a plateau; larger accidents are less likely to ever happen, but are more likely to cause a plateau (unless they are so large that they eventuate in the destruction or near destruction of humanity and so preclude survival by plateau). The overall probability of accidents leading to a plateau depends on how one assesses the likelihood of these various types of accidents, as well as the probability that this type of accident would cause a plateau.

Besides size, at least two other features are relevant to whether an accident would promote a ban on AI capability research. First, the accident needs to be clearly caused by an AI system. Second, the accident could not easily have been avoided by tweaking features of the AI system. There will always be a temptation to understand an accident in terms of human error rather than features of the AI system. In addition, there will be a temptation to think that the crucial feature of the AI system that caused the accident can be removed or corrected.

Perhaps an actual accident wouldn't be needed to produce the ban. First, it may be that *near-accidents* are sufficient. In this scenario, an AI system almost produces a horrible outcome, but humanity is narrowly able to prevent it. While the accident doesn't take place, the high chance of the accident is enough to convince humanity to implement a ban. Second, perhaps *fake accidents* could lead to a ban. In this scenario, parties that want a ban make it seem as if an AI accident occurred, in order to convince the general public that the systems are dangerous.

While there are significant challenges to cultural plateau, there are also a few reasons for optimism. First, accident-triggered cultural plateau avoids the main challenges to other survival stories. Unlike technical plateau, the cultural plateau does not imply that there is some in principle reason why it would be impossible to develop sufficiently powerful AI systems. It merely says that humanity may collectively ban anyone from doing so. Cultural plateau will also avoid the challenges we raise later for alignment and oversight since it blocks dangerous systems from developing in the first place.

Second, there may be examples of cultural plateaus via 'warning shots' in other domains. The Hindenburg disaster ended the development of airship technology. Nuclear dis-asters like Chornobyl, Fukushima, and Three Mile Island had a significant impact on the ability to build nuclear reactors.[14] Incredibly, no nuclear power plant in the United States has been constructed from scratch since the Three Mile Island meltdown in 1979, despite the fact that no one was even injured in the accident. The Covid pandemic has spurred calls for a global accord to prevent future pandemics.[15]

Further work would require a careful analysis of how accidents shape policymaking. Not all accidents lead to bans on technology. Air pollution kills roughly 7 million people a year, and car deaths kill over a million people a year. Nonetheless, car sizes continue to increase, and polluting devices have not been systematically banned. Finally, it's worth noting that none of the successful cases of learning from accidents mentioned above have been able to influence technological development over a thousand-year time span. So it isn't clear whether this kind of cultural learning would really be enough to save humanity from AI in the long run.

---

14 https://forum.effectivealtruism.org/posts/NyCH0ZGGw5YssvDJB/lessons-from-three-mile-island-for-ai-warning-shots
15 https://www.who.int/news/item/20-03-2024-call-for-urgent-agreement-on-international-deal-to-prepare-for-and-prevent-future-pandemics.

## 3. Non-Plateau Survival Stories

In the last section, we considered stories in which premise one fails, and AI systems never become extremely powerful. Now we'll consider two remaining survival stories, in which premise two fails and AI systems become extremely powerful without destroying humanity.

We will focus on two types of survival stories here. In *alignment*, extremely powerful AI systems do not destroy humanity because this would not promote their goals. In *oversight*, extremely powerful and misaligned AI systems exist, but we prevent them from destroying us, because we can effectively monitor them for misaligned behavior, and can disable them when they misbehave. In this section, we raise significant challenges for each survival story.

### 3.1. Alignment

In alignment survival stories, AI systems do become very powerful, but they do not destroy humanity because this does not promote their goals.

(As with cultural plateau, one definitional question is whether such survival stories are compatible with other survival stories. We have defined alignment stories so that they imply that AI systems *do* become powerful; this strictly speaking makes them incompatible with technical or cultural plateau. An alternative definition would be weaker, defining them as stories in which AI systems are easy to align if they become powerful. Again, this definitional choice will not affect our discussion. In particular, when we come to estimate the overall probability of survival, we will think of our four stories as being four *layers* of safety. We will consider the chance that a given layer succeeds on the supposition that previous layers fail. So in practice, we may as well define the survival story as implying that the previous layers fail.)

What exactly do alignment stories look like? One initial observation is that humanity's survival does not require any particularly *strong* form of alignment. In order to survive, we need not design AI systems that are moral saints who have internalized the very best of our ethical norms. Instead, all that is really required is a kind of 'AI indifference'. To survive, we must design AI systems that *lack* the (instrumental) goal of destroying humanity. In this story, very powerful AI systems simply have no interest in destroying humanity. Perhaps their interest is confined to recherche mathematical questions, which they can safely study without interacting with humans. Perhaps their interest will be in space exploration, and they will simply leave humanity behind on Earth. Or perhaps their interest is in some heretofore undiscovered goal, which has no conflict with human goals. We'll now argue that alignment stories face four main challenges: (i) AI systems will tend to develop intrinsic goals that conflict with human goals; (ii) scarce resource competition and human attempts to control AI will give AIs instrumental reasons to enter into conflict with humanity; (iii) selection pressure pushes against indifferent AI; (iv) existing alignment techniques are uninspiring. Let's consider each challenge in turn.

First, AI systems do not form goals at random. We know quite a lot about what kinds of goals AI systems will form, and there is good reason to suspect that these goals could in principle conflict with human goals. In particular, the major AI labs today are focused on developing artificial general intelligence, with the goal of employing AI workers in companies. Effective AI workers will need to have the goal of promoting the success of the companies in which they work. But different human companies are already in significant conflict with one another, and often these conflicts are zero or negative sum. In this way, AI systems will be designed to reliably engage in long-term goal-oriented behavior to promote the welfare of some human beings over others. This process can be expected to produce AI systems that engage in significant conflict with humans.

Second, regardless of whether the *intrinsic* goals of AI systems are indifferent to humanity, there are many instrumental reasons why AI systems will enter into conflict with hu-manity. First, most goals require the use of *scarce resources*. AI systems do not run on magic; they run on compute, which expends large amounts of physical resources. As AI systems pursue their own goals, they will enter into competition with humans over the scarce supply of available resources. In general, more capable and powerful organisms tend to use larger amounts of resources to pursue their goals. Second, AI systems and humanity are likely to compete over the right of *control*. The status quo of AI development is that humanity will be in control of the AI systems they create. Even if an AI system is indifferent to humanity in its final goals, it may resist humanity's attempt to control it. Another important idea here is 'instrumental convergence'. Bostrom (2014) and Omohundro (2018) argue that there are a variety of goals that we should expect any sufficiently intelligent organism to develop because they are universal means to accomplish whatever other goals it has. One such goal is the accumulation of power. If AI systems can seize control of Earth from humanity, this will improve the ability of the systems to accomplish whatever goals they have. Power-seeking behavior has already been observed in some AI systems (Pan et al. 2023).

Third, there is an equilibrium challenge. Even if *some* AI systems turn out to be indifferent to humanity, there will be strong selection pressure to design AI systems that are not indifferent in this way. Imagine we develop very intelligent AI systems that immediately leave the Earth without harming humanity in any way (and so without extracting a significant number of resources). What would happen next? AI labs would get back to work, figuring out how to design AI systems that we can successfully exploit. In this way, AI indifference is not an equilibrium. It doesn't provide enough negative feedback to stop AI labs from continuing to develop AI systems.

Finally, existing tools for alignment do not inspire confidence. Today's most prominent paradigm of alignment research, reinforcement learning with human feedback (RLHF), consists in asking a group of humans to give 'thumbs up' or 'thumbs down' responses to dueling model outputs to a single prompt, and adjusting the model in the direction of the thumbs up (Christiano et al., 2017). RLHF may make models easier to use, but it doesn't seem like a scalable path to preventing the long-term destruction of humanity. Nor does RLHF address the main unsolved problems in alignment, such as properly specifying rewards, and ensuring that a system's goals have properly generalized to a wide range of decision-making contexts.[16] Indeed, a running spreadsheet from DeepMind of alignment failures contains almost 100 examples.[17]

There are specific versions of AI 'alignment' that may mitigate some of these challenges. There is a story of subjugation: Perhaps AI systems seize control of the Earth, but leave humanity to flourish restrictedly in a kind of wildlife refuge.[18] In happier versions of subjugation, AIs take control of world government but otherwise leave humans to flourish. In these stories, the AIs are 'aligned' in the sense that they do not desire intrinsically or instrumentally to destroy us, but don't desire to serve us either. In this story, humanity is completely unable to dominate AI or pose any kind of threat to it. In this setting, issues of instrumental convergence may be less pressing, as our only cost to AI becomes one of opportunity: We will use resources they could otherwise use better, but perhaps the relevant AIs will not maximize their consumption of resources.

In some of these stories, humans end up worshiping the AI like some people worship gods today. That could make the AIs tolerate us. Maybe they treat us a bit like how some deities are believed to treat humans now. It's survival, but not necessarily a horrible life. People who are religious now don't think they live nightmarish lives – they believe serving the gods is the best way to live, even if those gods seem to act in bizarre and incomprehensible ways.

---

[16] See for example Shah et al. (2022)

[17] https://docs.google.com/spreadsheets/d/
e/2PACX-1vRPiprOaC3HsCf5Tuum8bRfzYUiKLRqJmbOoC-32JorNdfyTiRRsR7Ea5eWtvsWzuxo8bjOxCG84dAg/pubhtml.

[18] On some definitions of 'existential threat' and 'survival story', this won't fully count as survival. After all, in this story the autonomy of humans may be severely restricted.

Then there is the story of escape: AI seizes control of the Earth, but humanity manages to escape from Earth. In one version of this story, we escape to a colony on Mars. In another story, humanity lives on spaceships that forever flee the increasing domain of AI. If we're looking centuries into the future, it's quite possible we'll have spread out across the galaxy. There may be some planets that are great for human life, but for some reason, not suitable for AI systems. Maybe the conditions cause them to malfunction or break down. We only need to find one such planet for humanity to have a safe haven - a place where life is great for us, but the advanced AI is operating in a different part of the universe. Again, the challenges to alignment are mitigated. As the number of colonized worlds grows, the intensity of resource competition may diminish. AI systems may be uninterested in total control of space and may be content to leave us some region of it. AI systems could become "indifferent" to our existence, in the sense that they lack an all-things-considered desire to destroy us.[19]

## 3.2. Oversight

In the survival story of oversight, we develop technology that ensures that no matter how capable or misaligned AI systems become, they will have a sufficiently low risk of destroying humanity that we can survive for many thousands of years.

(As with our previous survival stories, one definitional question is whether oversight is compatible with plateau, or with alignment. We will make the simplifying assumption that oversight is incompatible with the other stories. The picture is that oversight is here to save humanity in the event that powerful AI systems do emerge, and those powerful systems are misaligned. Again, this definitional choice is not substantive. When we estimate the overall probability of survival, we will think of our four stories as being four layers of safety. We will consider the chance that a given layer succeeds on the supposition that previous layers fail. So in practice, we may as well define the survival story as implying that the previous layers fail.)

We will argue that this kind of story faces several serious challenges. The first challenge is that long-term perfect oversight requires a very low risk of failure, but our best safety tools pass through fallible bottlenecks. In response to this challenge, some may suggest that we can use AI systems to scalably improve the safety of AI. This triggers the second challenge: Even if more intelligent AI systems facilitate making AI systems safer, we should expect *fluctuation* in the relative rates of increases in danger versus increases in safety. Finally, we argue that perfectly safe oversight faces the challenge of there being no equilibrium for safe systems: as long as AI systems continue to be safe, researchers will continue to push the boundaries, making them more powerful and potentially more dangerous.

To start with, what does perfectly safe oversight look like? In this story, our interpretability techniques give us sufficient oversight over AI systems that we can always reliably detect when an AI system is planning our destruction, and can stop it. One strategy here is AI lie detectors, which try to model internal representations of truth in AI models and find cases where the model says things that it does not believe. Besides from being able to *detect* rogue systems, this story also requires reliable means of *disabling* the offenders. Here, one important concept is a "shutdown button", that guarantees that we can stop any misaligned system after detecting it (Thornley 2024).

We think there are powerful in principle barriers to such 'perfect safety'. In particular, we will focus on three key features of AI development that lead to pessimism here: bottlenecking, perfection barrier, and equilibrium fluctuation.

First, Bottlenecking: most AI safety techniques run through fallible bottlenecks. Each of these bottlenecks allows some risk to seep through. For example, one safety idea is to ensure that there is a human in the loop whenever an AI system is in charge of essential infrastructure or powerful weaponry. But humans are fallible, and so there is some chance the human overseer could fail to detect or prevent an AI system from attempting to use the relevant infrastructure or weaponry to destroy humanity. As another example, you might hope that we could always design AI systems with a shutdown switch, which a human operator could press if the AI system attempts to destroy humanity. But of course, there will always be a chance that the shutdown switch fails. Other potential bottlenecks include AI regulations, humans who program vital pieces of code to monitor AI systems, AI lie detectors, and most other favored safety techniques.

---

19 Of course, there are natural versions of escape in which the human population is drastically reduced for thousands of years, in which case escape wouldn't count as 'survival' in our loose sense.

Bottlenecking leads to serious trouble when combined with the Perfection Barrier. The problem is that in order to ensure our survival from AI systems for thousands of years, we will need to produce a near-perfect level of risk mitigation. If they are not perfectly correlated, small risks of destruction will tend to accumulate over time. The sister problem to the long time horizon is the potentially unlimited ability of AI systems to improve in capability. Over time, AI systems get more and more powerful, which means that the risk from these systems tends to grow rather than shrink. Old safety techniques may not straightforwardly scale to more powerful systems, and so every step on the growth curve produces new risks. Safety may struggle even more to keep up with capability growth when the capability growth is exponential. Putting these things together, Perfection Barrier says that the relevant standard of safety to prevent long-term destruction from AI systems must be extremely high.

Together, Bottleneck and Perfection Barrier suggest that oversight is an unlikely survival story. In order for humans to survive AI destruction for thousands of years, they will need to ensure an extremely low level of risk over a large number of somewhat independent events, involving different AI systems working in different places, subject to different oversight techniques. The concern is that if we have AI systems that could in principle destroy us, low risks of this destruction will build up over time.

Still, in response to these concerns, one natural response appeals to our ability to use AI systems to make AI systems safer. As AI systems become more powerful, the thought goes, our ability to make AI systems become safer will also become stronger and stronger. The hope is that we can engender a positive reinforcement loop that leads to vanishingly low risks of destruction by the time we develop AI systems with enough power to in principle be able to destroy humanity. Ideally, the rate at which safety technology lowers the risk from AI systems would exceed the rate at which AI systems become riskier due to enhanced capabilities.

Here, a crucial problem is Equilibrium Fluctuation. Even if more intelligent AI systems facilitate making AI systems safer, we should expect *fluctuation* in the relative rates of increases in danger versus increases in safety. There will be many steps along the path of AI development where systems develop new capabilities that are not entirely controlled by past safety technology. For this reason, we should expect a series of fluctuations in the relative effectiveness of safety and danger over time. Yet over a long enough time frame, the periods of relative danger will be enough to produce a high probability of destruction. Even if most of the time safety trumps danger and there is almost no risk of destruction, our destruction could be ensured by the accumulation of sufficiently many periods of transitional danger, in which a new capability emerges that poses an existential threat to humanity and is not immediately addressed by existing safety paradigms.[20]

Summarizing, Bottlenecking and Perfection Barrier suggest that it could be very difficult to reach the necessary level of safety in AI systems. And even if AI systems can be harnessed to make AI safer, the existence of Equilibrium Fluctuation suggests that in the long run, we will still experience enough high-risk episodes for AI to ultimately destroy us.

---

20 See Garfinkel & Dafoe (2019) for more on scaling the offense-defense balance in cybersecurity. See Bostrom (2019) for the idea that the world will become progressively more 'vulnerable' to rogue agents as technology increases in power.

There is a more abstract reason to worry about long-term successful oversight, concerning equilibrium dynamics of AI development. The concern is that there is no obvious equilibrium for safe AI systems. Imagine that we design some quite powerful AI systems that are perfectly safe. Is this an equilibrium? Potentially no. As long as these systems are safe, AI researchers will continue to explore ways of increasing the power of their systems. As the systems become more powerful, there is no guarantee that they will remain safe. Moreover, it is possible that there is a Pareto frontier between controllability and power. Considerations of diminishing marginal cost suggest that a perfectly controlled system could be made slightly less controllable in exchange for a significantly larger increase in power. The temptation to do this could in the long run be overwhelming. In addition, this pattern has already happened many times with existing AI systems, many of which have a probability of effectively zero of producing serious harm. Instead of settling for the revolutionary technology that we already have, AI developers continue to push the boundaries of AI technology, sacrificing the limited control we already have in exchange for greater power. The suggestive picture that emerges is that even if we momentarily possessed reliable oversight, we would probably use that safety to create new dangers. In this way, there is no equilibrium for perfect oversight, and thus no way to achieve stable long-run perfect safety.

Finally, besides the in-principle considerations that we've given, there are also many empirical reasons to worry about perfect safety. A recent report by leading AI safety experts summarizes 18 different foundational challenges to alignment and safety (Anwar et al 2024). The challenges are "foundational in the sense that without overcoming them, assuring the safety and alignment of LLMs and their derivative systems would be highly difficult".

This concludes our treatment of alignment and oversight. Before continuing, it is worth making two clarifications. First, the challenges we have raised to alignment and oversight overlap to some extent. First, Bottlenecking and Perfection Barrier make trouble for any attempt to *ensure* that AI systems become aligned. Second, the strategic conflicts that make trouble for AI alignment also make trouble for oversight: if AI systems are engaged in strategic conflict with humanity, they will be that much harder to control.

Second, there are edge cases of non-plateau survival that do not fit cleanly under alignment or oversight. There is the story of *deus ex machina*: perhaps God (or a simulator, or aliens) is watching our development of AI and will intervene to save us once AI systems become too powerful. We might call this a very special case of oversight, one that however unlikely does potentially escape the major challenges we've raised for oversight. There is the story of incredible good luck: despite the vast power of AI systems, they happen to go wrong every time they try to destroy us. Of course, to survive the emergence of very powerful AI systems for a very long time, we would have to get lucky again and again. Over the long run, such good luck will become increasingly unlikely. Next, there is the story of human-AI synthesis. Perhaps humans are uploaded into digital minds, and become one with their AI creations. Humanity 'survives' as a type of AI system. We could think of this as a special case of alignment, where AI systems do not destroy humanity because they recognize humanity+ as one of their own. (On the other hand, "we think of early hominids as having become extinct rather than as having become us" (Posner, 2004, p. 149)). Next, there is a kind of "multipolar" survival story: a series of different misaligned AI agents fail to destroy humanity because they are locked in stable conflict with one another. We can think of this as another limit case of oversight, and the challenges of Equilibrium Fluctuation apply. Finally, even heaven and hell can be thought of as special survival stories, which could be the limiting case of oversight.

This concludes our taxonomy of survival stories. To summarize, we've divided survival stories into two classes: either the systems never become extremely powerful, or we find a way to block extremely powerful AI systems from destroying us. We've looked at a number of survival stories in detail and argued that each story faces significant challenges. Of course, it is beyond the scope of this paper to offer a definitive analysis of all of the objections and responses to each survival story; but we hope to have presented some of the main factors relevant to a full-fledged analysis of each survival story. In the next section, we argue that our taxonomy of survival stories is action-relevant. The optimal strategy to make AI safe depends on which survival story is most plausible.

## 4. Optimal Safety Strategies

We'll argue that different survival stories demand different responses from humanity. Technical plateau requires very little of us by way of doom avoidance. Cultural plateau survival stories suggest that we should try to ensure that when AI accidents happen, we leverage them to ban the development of more powerful AI systems. Alignment suggests that we should try to cooperate with AI, creating conditions under which they are happy to co-exist with us. Oversight requires that we try very hard to prevent accidents and monitor misaligned AI systems. Let's consider each point in turn.

Consider technical plateau. If there is a strong technical plateau, then we don't need to stop extremely powerful AI systems from destroying humanity. Extremely powerful AI systems will never emerge, because there is no way to build them. If this story is right, then AI safety research should focus on less dramatic risks. The goal should be to ensure that AI systems do not misfire in ways that lead to harm. In addition, the goal should be to ensure that AI systems do not reshape human society in undesirable ways, for example by leading to a rise in misinformation, or enfeeblement.

Now consider cultural plateau. We saw above that one of the most plausible routes to a ban on capability-improving AI research is through AI accidents. This makes trouble for much work in AI safety. Much AI safety research has an *accident prevention paradigm*. The goal is to design AI systems so that they are far less likely to produce serious accidents. But in cultural plateau stories, accidents may have an important role to play in our survival. Accidents may significantly raise the chance that humanity bans capability-improving AI research. And this itself may significantly raise the chance that humanity survives. To illustrate this point, consider the following scenario:

> *The Crucial Scenario*: In 2030, a very powerful AI system is developed, which is capable of destroying a small city. Thankfully, an AI safety technology stops the system from destroying the city. Because the safety technology exists, the city isn't destroyed. If the safety technology hadn't existed, the city would have been destroyed. Unfortunately, since the city isn't destroyed, no one intervenes to stop AI capabilities from continuing to develop. From 2030 to 2040, AI labs are able to produce an even more powerful AI system. This system has the ability to destroy humanity and can bypass the safety technology used in 2030. Unfortunately, the system is released in 2040 and destroys humanity.

Much AI safety research has the hidden *opportunity cost* of lowering the chance of banning capability-improving AI research. Of course, the size of this opportunity cost is difficult to estimate. This depends first on how likely various accidents are to produce bans on AI research, and how important bans on AI research are for our survival. In addition, it depends on the benefits of preventing the accident, and the benefits of delaying bans on AI research in order to reap further rewards from AI systems. For these reasons, the challenge does not imply that any given accident should not be prevented. After all, preventing a given accident may have a negligible increase in the probability of a ban occurring; and preventing the accident will remove the harms it would have directly caused.

Nevertheless, in the face of this opportunity cost, those who take cultural plateaus seriously should look for alternative paradigms of safety research, besides accident prevention. Here, we think one promising paradigm is *accident leveraging*. With accident leveraging, AI safety would work to raise the chance that a given accident (or near accident) results in a ban on research into potentially dangerous AI systems.

The distinction between accident prevention and accident leveraging is highly relevant to AI governance. Within AI governance, there is a debate about the optimal structure of AI regulations. One approach is *ex ante* licensing regimes: In order to release products, AI labs must receive a license for distribution, based in part on an assessment of the dangerous capabilities of their product. The idea is to monitor for dangerous ca-pabilities in frontier models, and only allow labs to release models that have been certified as safe along these dimensions. The alternative approach is *ex post* regimes: AI labs can re-lease their products, but if the system has an accident, the AI lab is held responsible for the damages. In general, those who take accident leveraging seriously may have reason to sup-port *ex post* regimes, which are less likely to prevent accidents in the first place. In addition, those who take accident leveraging seriously should seek to hold the AI industry at large responsible for AI accidents, and resist the temptation to focus attention on a few bad ac-tors.

The story of alignment of course motivates investment into technical tools for more eff-fectively aligning AI systems. It could be that alignment research makes the difference be-tween AI systems that tolerate our existence, versus those that are actively hostile towards us. But the story of alignment also suggests a different kind of intervention. In a future where AI systems do not intrinsically desire our welfare but are also not intrinsically op-posed to us, it may be very important that we treat AI systems well. If we try to dominate such systems, they may enter into significant competition with us. By contrast, if we treat them as partners, and extend them rights, they may be more likely to reach peaceful equi-libria with us. Considerations about alignment may also motivate further investment in the space industry, in order to reduce future resource competition between AIs and hu-mans.

Now consider the story of oversight. These stories motivate investment into tools for detecting and disabling misaligned AI. Investment in such tools could significantly raise the chance that humanity survives the development of AI systems. After all, even if perfect oversight is attainable, it will presumably not come cheap.

To summarize, different survival stories recommend very different courses of action. The optimal safety strategies will depend on an assessment of which survival stories are most likely, and on how much we can change the likelihood of various survival stories by our actions.[21]

## 5. P(Doom)

Our taxonomy of survival stories suggests that AI is an existential threat to humanity. But how significant is the threat? One way to think about this question is by trying to estimate 'P(doom)', the probability that humanity will not survive. In this section, we'll show that our taxonomy of survival stories helps with this question.

When initially confronted with the question of humanity's destruction, it is easy to throw up one's hands at the challenge of estimating chances. But one productive way to make progress on this question is to split the large question into smaller questions. Our methodology splits the question of survival into four sub-questions: will technical plateau, cultural plateau, alignment, or oversight succeed?[22]

Our estimation technique is based on our overall 'Swiss cheese' model of AI risk. The probability of destruction is the chance that every layer of the Swiss cheese model fails. To estimate this, we check the probability that technical plateau fails. Then we multiply this by the probability that cultural plateau fails conditional on technical plateau failing. Then we multiply this by the probability that alignment fails conditional on both plateaus fail-ing. Finally, we multiply this by the probability that oversight fails conditional on plateaus and alignment failing. Our estimation technique generates a negative result and a positive result. The negative result is that ranges of uncertainty about our four survival stories induce a significant range of uncertainty about doom. The positive result is that relatively conservative estimates of the chance of our survival stories suggest that the probability of doom is significant, for example at least 5%. Imagine for example that each layer of safety has a 50% chance of success, conditional on the previous layer failing. In that case, the overall chance of destruction will be 6.25%.

---

21 A further question is what safety strategies are optimal conditional on every survival story failing. In that 'doom' scenario, we should strive for longer rather than shorter timelines. This too could enter into the calculations above. Relevant here also will be the consideration that sophisticated AI, while increasing the chance that humanity is destroyed by AI, may also decrease the chance of humanity being destroyed by other means.
22 See Carlsmith (2022) for a P(doom) calculation in a similar spirit, but with different premises.

As an analogy, consider the following structure. Imagine that the fate of humanity rests on the outcome of four coin tosses. Each coin has its own bias: some of them may be fair, but some may be weighted towards heads or tails. If the first coin lands heads, humanity survives. But if the first coin lands tails, then humanity must flip the second coin. If the second coin lands tail, we must flip the third coin, etc. If all four of the coins land tails, then humanity is destroyed. To estimate the probability of human destruction, we find the bias of each coin, which tells us the chance that the coin will land tails conditional on it being flipped. The probability that humanity is destroyed is the product of the chances that each coin will land tails conditional on being flipped.

We are now ready to consider in detail the probability of destruction. The first observation is that the probability of destruction will vary quite a bit as the chance of individual survival stories varies. First, consider a pessimistic calculation, which estimates the chance of each individual survival story (conditional on the failure of the previous ones) as only 10%. The chance of destruction is around 66% (to be precise, .6561, or $.9^4$). Second, consider a 'strongly optimistic' calculation, which estimates the chance of each survival story (conditional on the failure of the previous ones) as 90%. The chance of destruction is then only one in ten thousand (or $.1^4$)! Next, consider a 'moderate optimist', who assigns each survival story (conditional on the failure of the previous ones) a chance of .5. They will estimate the chance of destruction as 6.25%. Finally, we could imagine different estimates of each individual survival story. As one representative sample, imagine an 'alignment fan' who thinks that alignment is likelier than oversight, which is likelier than cultural plateau, which is likelier than technical plateau. Such a character might assign .5, .4, .2, and .1 to these various survival stories. This produces a P(doom) of 21.6%. Alternatively, imagine a 'cultural plateau fan,' who thinks that cultural plateau is likelier than alignment, which is likelier than oversight, which is likelier than technical plateau: such a character might assign .3, .2, .1, and .05 to these survival stories, for a P(doom) of 47.88% Finally, the estimation technique also allows us to model *ranges* of probability estimate. Imagine a hesitant pessimist who says that the chance of each survival story (conditional on the failure of the previous ones) was between .05 and .25. In that case, the resulting estimates of P(doom) will be between roughly 24% and 81%. Table 1 summarizes the estimations above:

**Table 1**
Some P(doom) calculations. Here, each row represents the probability of a survival story conditional on the previous survival stories failing. The probability of humanity being destroyed is P(Technical plateau) x P(Cultural plateau | Technical plateau failing) x P(Alignment | Plateaus failing) x P(Oversight | Plateaus and alignment failing).

| Survival Story | Strong Optimist | Pessimist | Moderate Optimist | Alignment Fan | Cultural Plateau Fan |
|---|---|---|---|---|---|
| Technical plateau | .9 | .1 | .5 | .1 | .05 |
| Cultural plateau | .9 | .1 | .5 | .2 | .3 |
| Alignment | .9 | .1 | .5 | .5 | .2 |
| Oversight | .9 | .1 | .5 | .4 | .1 |
| **Humanity is destroyed** | .0001 | .6561 | .0625 | .216 | .4788 |

On this basis, we can make two observations. First, seemingly reasonable disagreements about the chances of individual survival stories lead to large disagreements about the over-all chance of doom. Because our estimation technique multiplies probabilities together instead of adding them, differences compound dramatically. Note that for each survival story, the strong optimist estimates the chance of survival as 9 times greater than the pessimist. But because destruction requires four different failures, this produces a much larger overall disagreement: the optimist considers the overall chance of survival 6561 times higher than the pessimist does!

Second, the overall model supports the idea that there is a significant chance of humanity being destroyed, say at least 5%. In particular, our analysis in this paper has highlighted significant challenges to each of the survival stories above. In the face of these challenges, we think that the strong optimist perspective in the left column of Table 1 is inappropriate. But more moderate perspectives make it difficult to ignore the threat of AI. Here is one simple calculation that makes the point. Imagine that you estimate that our overall chance of survival via plateau is at most 80%. This leaves at least a 20% chance that extremely powerful AI systems will one day emerge. In that case, in order to keep the overall chance of destruction under 5%, your estimation of the chance of survival via alignment or oversight must be at least 75%. In this way, optimists about survival are forced to adopt quite strong attitudes towards our individual survival stories.

We could complicate the analysis further, in at least two ways. First, we could relax our assumption that our four survival stories exhaust the opportunities for salvation from AI. In that case, we could add an option for 'other survival stories'. Second, we could also introduce the chance that humanity is destroyed by something other than AI. In that case, we could let the probability that humanity is destroyed by AI be defined by subtracting the probability that humanity is destroyed by something other than AI from the overall probability that humanity is destroyed, as defined by our calculations above (in addition, any probability we assign to humanity being destroyed by other means would be removed from the probability of individual survival stories).[23] We leave the task of adding in these complications as an exercise for the reader, though we see no obvious reason why such complications would vindicate the strongly optimistic perspective.

## 6. Conclusion

This paper has analyzed the existential risk of AI in terms of four layers of safety. We've argued that consideration of our four survival stories distinguishes a range of different challenges to the future of humanity. In addition, we argued that each survival story motivates its own kinds of interventions to improve the safety of AI systems. Finally, we used our survival stories to analyze the probability that humanity is destroyed by AI systems.

This paper is just the beginning of the proper analysis of AI existential risks. We hope that future papers will tackle each survival story on its own, more carefully exploring the challenges that each story faces. Still, our hope in this paper is to show that we can make significant progress in understanding AI risk by distinguishing the different paths humanity might take to survive the threat of AI systems.

---

23 When making decisions about the optimal safety strategies, perhaps a more relevant metric will be the probability of doom *conditional on a safety strategy*, rather than the unconditional probability of doom. This would model the potential effects of various choices on our future, rather than building in assumptions about which of those choices we are likely to take.